\documentclass[11pt,a4paper]{article}
\usepackage[dvipsnames]{xcolor}
\usepackage[hyperref]{emnlp-ijcnlp-2019}
\usepackage{times}
\usepackage{latexsym}

\usepackage{url}
\usepackage{amsmath}
\usepackage{multirow}
\usepackage{arydshln}
\usepackage{graphicx}
\usepackage{subcaption}
\usepackage{comment}
\usepackage{soul}
\usepackage{setspace}
\usepackage{anyfontsize}
\usepackage{xspace}

\aclfinalcopy % Uncomment this line for the final submission

\newcommand{\dataset}{BASIL\xspace}

\sethlcolor{orange}

\title{In Plain Sight: Media Bias through the Lens of Factual Reporting}

\author{Lisa Fan$^{1,}$\thanks{Equal contribution. Lisa Fan focused on annotation schema design and writing, Marshall White focused on data collection and statistical analysis.}\hspace{.75cm} Marshall White$^{1,*}$\hspace{.75cm} Eva Sharma$^1$\hspace{.75cm} Ruisi Su$^1$\hspace{.75cm}\\{\bf Prafulla Kumar Choubey$^2$ \hspace{.75cm} Ruihong Huang$^2$ \hspace{.75cm} Lu Wang$^1$}\\
$^1$ Khoury College of Computer Sciences, Northeastern University, Boston, MA 02115 \\
$^2$ Department of Computer Science and Engineering, Texas A\&M University\\
{ \tt \{fan.lis, white.mars, sharma.ev, su.ruis\}@husky.neu.edu} \\
{\tt \{prafulla.choubey, huangrh\}@tamu.edu, luwang@ccs.neu.edu}
}

\date{}

\begin{document}
\maketitle

\begin{abstract}
The increasing prevalence of political bias in news media calls for greater public awareness of it, as well as robust methods for its detection.
While prior work in NLP has primarily focused on the \textit{lexical bias} captured by linguistic attributes such as word choice and syntax, other types of bias stem from the actual content selected for inclusion in the text.
In this work, we investigate the effects of \textit{informational bias}: factual content that can nevertheless be deployed to sway reader opinion. 
We first produce a new dataset, \textsc{\dataset}, of 300 news articles annotated with 1,727 bias spans\footnote{Dataset can be found at \url{www.ccs.neu.edu/home/luwang/data.html}.} and find evidence that informational bias appears in news articles more frequently than lexical bias.
We further study our annotations to observe how informational bias surfaces in news articles by different media outlets.
Lastly, a baseline model for informational bias prediction is presented by fine-tuning BERT on our labeled data, indicating the challenges of the task and future directions. 
\end{abstract}

\section{Introduction}\label{intro}
% \tbd{OVERALL: check verb tense, transitions}
News media exercises the vast power of swaying public opinion through the way it selects and crafts information \citep{de2004effects,dellavigna2010persuasion,mccombs2009news,perse2001media,reynolds2002news}. Multiple studies have identified the correlation between the increasing polarization of media
and the general population's political stance
% and the increasing polarization of our world at large
\citep{gentzkow2010drives,gentzkow2011ideological,prior2013media},
underscoring the imperative to understand the nature of news bias and how to accurately detect it.

\newcommand{\infbias}[1]{\textit{#1}}
\newcommand{\lexbias}[1]{\textbf{#1}}

\begin{figure}[t]
\centering
% \small
% \footnotesize
\fontsize{8.5}{10.83}
% \fontfamily{qhv}\selectfont
\fontfamily{phv}\selectfont
\begin{tabular}{|p{.94\linewidth}|}
\hline
\textbf{Main event:} Democratic presidential candidates ask to see full Mueller report\\
\textbf{Main targets:} \textcolor{red}{Donald Trump}, \textcolor{blue}{Democratic candidates}\\
\hline
\textbf{HPO:} Democrats want access to special counsel Robert Mueller's investigation into Russian interference in the 2016 presidential election [\infbias{before President Donald Trump has a chance to interfere.}]$_{\textcolor{red}{\text{Trump}}}$ ... Sen. Mark Warner said in a statement: [\infbias{``Any attempt by the \underline{Trump Administration} to cover up the results of this investigation into Russia's attack on our democracy would be unacceptable.''}]$_{\textcolor{red}{\text{Trump}}}$\\
\hdashline
\textbf{FOX:} Democratic presidential candidates [\lexbias{wasted no time}]$_{\textcolor{blue}{\text{Dems}}}$ Friday evening demanding the immediate public release of the long-awaited report from Robert S. Mueller III. ... Several candidates, in calling for the swift release of the report, also [\infbias{sought to gather new supporters and their email addresses}]$_{\textcolor{blue}{\text{Dems}}}$ by putting out [\lexbias{``petitions''}]$_{\textcolor{blue}{\text{Dems}}}$ calling for complete transparency from the Justice Department.\\
\hdashline
\textbf{NYT:} And on Saturday, one day before Attorney General William Barr released a short summary of Mueller's findings, former Texas Rep. Beto O'Rourke charged on the campaign trail in South Carolina that you [\infbias{``have a president, who in my opinion beyond the shadow of a doubt, sought to, however} [\infbias{\lexbias{ham-handedly,}}]$_{\textcolor{red}{\text{Trump}}}$\infbias{ collude with the Russian government---a foreign power---to undermine and influence our elections.''}]$_{\textcolor{red}{\text{Trump}}}$\\
\hline
\end{tabular}
\vspace{-2mm}
\caption{
%\lu{fig too big, see right}
Examples of negative bias from Huffington Post (HPO), Fox News (FOX), and New York Times (NYT) discussing the same event. \textit{Informational bias} and \textbf{lexical bias} are highlighted. The target of the bias is noted at the end of each span. Intermediary targets of indirect bias spans are \underline{underlined}.}\label{fig:example}
\vspace{-3mm}
\end{figure}

% HPO: second annotation is an example of indirect bias (main target is Trump, indirect target is Trump Administration). FOX: how framing works even on the same topic. NYT: how articles use heavily opinionated quotes
In the natural language processing community, the study of bias has centered around what we term \textbf{lexical bias:} bias stemming from content realization, or how things are said
\citep{greene2009more,hube2019neural,iyyer2014political,recasens2013linguistic,yano2010shedding}.
Such forms of bias typically do not depend on context outside of the sentence and can be alleviated while maintaining its semantics: polarized words can be removed or replaced, and clauses written in active voice can be rewritten in passive voice.

However, political science researchers find that news bias can also be characterized by decisions made regarding content selection and organization within articles \citep{gentzkow2015media,prat2013political}. 
As shown in Figure~\ref{fig:example}, though all three articles report on the same event, Huffington Post (HPO) and Fox News (FOX) each frame entities of opposing stances negatively: HPO states an assumed future action of {\it Donald Trump} as a fact, and FOX implies {\it Democrats} are taking advantage of political turmoil.
Such bias can only be revealed by gathering information from a variety of sources or by analyzing how an entity is covered throughout the article.

We define these types of bias as \textbf{informational bias:} sentences or clauses that convey information tangential, speculative, or as background to the main event in order to sway readers' opinions towards entities in the news.
Informational bias often depends on the broader context of an article, such as in the second FOX annotation in Figure~\ref{fig:example}: 
gathering new supporters would be benign in an article describing political campaign efforts. The subtlety of informational bias can more easily affect an unsuspecting reader, which presents the necessity of developing novel detection methods.

In order to study the differences between these two types of bias, we first collect and label a dataset, \textsc{\textbf{\dataset}} 
(\underline{B}ias \underline{A}nnotation \underline{S}pans on the \underline{I}nformational \underline{L}evel), of 300 news articles with lexical and informational bias spans. 
To examine how media sources encode bias differently, the dataset uses 100 triplets of articles, each reporting the same event from three outlets of different ideology. Based on our annotations, we find that all three sources use more informational bias than lexical bias, and informational bias is embedded uniformly across the entire article, while lexical bias is frequently observed at the beginning. 

We further explore the challenges in bias detection and benchmark \dataset using rule-based classifiers and the BERT model \citep{devlin2019bert} fine-tuned on our data. Results show that identifying informational bias poses additional difficulty and suggest future directions of encoding contextual knowledge from the full articles as well as reporting by other media. 

\section{Related Work}\label{related}

Prior work on automatic bias detection based on natural language processing methods primarily deals with finding sentence-level bias and considers linguistic attributes like word polarity \citep{recasens2013linguistic}, partisan phrases \citep{yano2010shedding}, and verb transitivity \citep{greene2009more}.
However, such studies fail to take into consideration biases that depend on a larger context, which is what we try to address in this work. 

Our work is also in line with \textit{framing analysis} in social science theory, or the concept of selecting and signifying specific aspects of an event to promote a particular interpretation \citep{entman1993framing}. 
In fact, informational bias can be considered a specific form of framing where the author intends to influence the reader's opinion of an entity. The relationship between framing and news is investigated by \citet{card2015media}, in which news articles are annotated with framing dimensions like ``legality'' and ``public opinion.'' \dataset contains richer information that allows us to study the purpose of ``frames,'' i.e., how biased content is invoked to support or oppose the issue at hand.

Research in political science has also studied bias induced by the inclusion or omission of certain facts \citep{entman2007framing,gentzkow2006media,gentzkow2010drives,prat2013political}. 
However, their definition of bias is typically grounded in how a reader perceives the ideological leaning of the article and news outlet, whereas our informational bias centers around the media's sentiment towards individual entities. 
Furthermore, while previous work mostly uses all articles published by a news outlet to estimate their ideology~\cite{budak2016fair}, we focus on stories of the same events reported by different outlets.

\begin{table*}
\centering
\small
\begin{tabular}{c}
% \begin{tabular}{|c c c|}
% \hline
% Articles & Sentences & Words \\
% \hline
% 300 & 7,984 & 224,163 \\
% \hline
% \hline
% Sents/Article & Words/Sent & Words/Article\\
% \hline
% 26.6 $\pm$ 12.2 &  28.1 $\pm$ 13.2 & 747.2 $\pm$ 343.6\\
% \hline
% \end{tabular}
\begin{tabular}{|c|c|c|c|c||c|}
\hline
\multicolumn{2}{|c|}{} &{\bf NYT} & {\bf FOX} & {\bf HPO} & {\bf All}\\
\hline
% NYT | FOX | HPO | all
\multicolumn{2}{|c|}{\# Articles} & 100 & 100 & 100 & 300 \\
\multicolumn{2}{|c|}{\# Sentences} & 3,049 & 2,639 & 2,296 & 7,984 \\
\multicolumn{2}{|c|}{\# Words} & 91,818 & 70,024 & 62,321 & 224,163\\
\multicolumn{2}{|c|}{\# Annotations} & 636 & 573 & 518 & 1,727 \\ 
\multicolumn{2}{|c|}{Sentences / Article} & 30.5 $\pm$ 13.8 & 26.4 $\pm$ 10.2 & 23.0 $\pm$ 11.0 & 26.6 $\pm$ 12.2 \\
\multicolumn{2}{|c|}{Words / Sentence} & 30.1 $\pm$ 14.0 & 26.5 $\pm$ 12.4 & 27.1 $\pm$ 12.5 & 28.1 $\pm$ 13.2  \\
% Words/Article & 918.2 $\pm$ 404.4 & 700.2 $\pm$ 257.8 & 623.2 $\pm$ 278.1 & 747.2 $\pm$ 343.6 \\
\multicolumn{2}{|c|}{Annotations / Article} & 6.4 $\pm$ 4.1 & 5.7 $\pm$ 3.8 & 5.2 $\pm$ 3.5 & 5.8 $\pm$ 3.8  \\
%\multicolumn{2}{|c|}{Words / Annotation} & 13.0 $\pm$ 11.6 & 13.0 $\pm$ 11.8 & 12.9 $\pm$ 12.0 & 13.0 $\pm$ 11.8\\ 
\hline
\multirow{2}{*}{{\bf Bias Type}} 
& {\it Informational} & 468 (73.6\%)  & 421 (73.5\%) & 360 (69.5\%)  & 1,249 (72.3\%)\\
% \cline{2-6}
& {\it Lexical}     & 168 (26.4\%) & 152 (26.5\%) & 158 (30.5\%)  & 478 (27.7\%) \\
\hline
\multirow{2}{*}{{\bf Aim}} 
& {\it Direct} & 574 (90.2\%) & 485 (84.6\%) & 462 (89.2\%) & 1,521 (88.1\%)\\ 
% \cline{2-6}
& {\it Indirect} & 62 (9.8\%) & 88 (15.4\%) & 56 (10.8\%) & 206 (11.9\%)\\
\hline
\multirow{2}{*}{{\bf Polarity}} 
& {\it Positive} & 112 (17.6\%) & 89 (15.5\%) & 110 (21.2\%) & 311 (18.0\%)\\ 
% \cline{2-6}
& {\it Negative} & 524 (82.4\%) & 484 (84.5\%) & 408 (78.8\%) & 1,416 (82.0\%)\\ 
\hline
\multicolumn{2}{|c|}{{\bf Annotations in quotes}} & 205 (32.2\%) & 299 (52.2\%) & 217 (41.9\%) & 721 (41.8\%) \\
\hline
\end{tabular}
\end{tabular}
\vspace{-2mm}
\caption{Descriptive statistics of the \textsc{\dataset} dataset. Mean and standard deviation shown where applicable. Annotation dimensions show raw counts and their percentage within the dimension in parentheses. %\lu{parentheses?} %\tbd{Note to Lu: Take out wrds/ann, since stat doesn't mean much if you don't separate inf/lex?}
}\label{tab:descriptive}
\vspace{-3mm}
\end{table*}

\section{\textsc{\dataset} Dataset Annotation}\label{schema}
% \section{Data}\label{data}

Using a combination of algorithmic alignment and manual inspection, we select 100 sets of articles, each set discussing the {\it same event} from three different news outlets. 10 sets are selected for each year from 2010 to 2019. 
We use, in order from most conservative to most liberal, Fox News (FOX), New York Times (NYT), and Huffington Post (HPO). {\bf Main events} and {\bf main entities} are manually identified for each article prior to annotation. {The political leanings of the main entities (\textit{liberal}, \textit{conservative}, or \textit{neutral}) are also manually annotated.} See the Supplementary for details. 

\smallskip
\noindent \textbf{Annotation Process.}
To compare how the three media sources discuss a story, annotators treat each article triplet as a single unit without knowing media information. 
Annotations are conducted on both {\it document-level} and {\it sentence-level}. On the document-level, annotators estimate the overall polarities of how the main event and main entities are covered, and rank the triplet's articles on the ideological spectrum with respect to one another. Before reading the articles, annotators specify their sentiment towards each main entity on a 5 point Likert scale.\footnote{The likely effect of annotators' prior beliefs on their perception of bias will be investigated in future work.}

On the sentence-level, annotators identify spans of lexical and informational bias by analyzing whether the text tends to affect a reader's feeling towards one of the main entities. 
In addition to the main dimension of {\bf bias type}
(\textit{lexical} or \textit{informational}), each span is labeled with the \textbf{target} of the bias (a \textit{main entity}), the bias \textbf{polarity} (\textit{positive} or \textit{negative} towards the target), the bias \textbf{aim} towards the main target (\textit{direct} or \textit{indirect}), and whether the bias is part of a \textbf{quote}.
Bias aim investigates the case where the main entity is indirectly targeted through an intermediary figure (see the HPO example in Figure~\ref{fig:example}, where the sentiment towards the intermediary entity ``Trump Administration'' is transferred to the main target, ``Donald Trump'').
Statistics are presented in Table~\ref{tab:descriptive}.

\smallskip
\noindent \textbf{Inter-annotator Agreement (IAA).}
Two annotators individually annotate each article triplet before discussing their annotations together to resolve conflicts and agree on ``gold-standard'' labels. 
We measure span-level agreement according to \citet{toprak2010sentence}, where we calculate the F1 score of span overlaps between two sets of annotations (details are in the Supplementary). 
Although the F1 scores of IAA are unsurprisingly low for this highly variable task, the score dramatically increases when agreement is calculated between individual annotations and the gold standard---from 0.34 to 0.70 for informational bias spans and from 0.14 to 0.56 for the sparser lexical spans, demonstrating the effectiveness of resolution discussions.

During the discussions, we noticed several trends that improved the quality of the gold standard annotations. First, the difficulty of being continually vigilant of one's own implicit bias would sometimes cause annotators to mark policies they disagreed with as negative bias (e.g., a liberal annotator might consider the detail that a politician supports an anti-abortion law as negative bias).
Discussions allowed annotators to re-examine the articles from a more neutral perspective. 
Annotators also disagreed on whether a detail was relevant background or biasing peripheral information. During discussions, they performed comparisons to other articles of the triplet to make a final decision---if another article includes the same information, it is likely relevant to the main event. This strategy reiterates the importance of leveraging different media sources.

For overlapping spans, we find high agreement on the other annotation dimensions, with an average Cohen's $\kappa$ of 0.84 for polarity and 0.92 for target main entity.

\section{Media Bias Analysis}
\begin{figure}
\centering
\resizebox{\linewidth}{!}{%
\includegraphics[height=5cm]{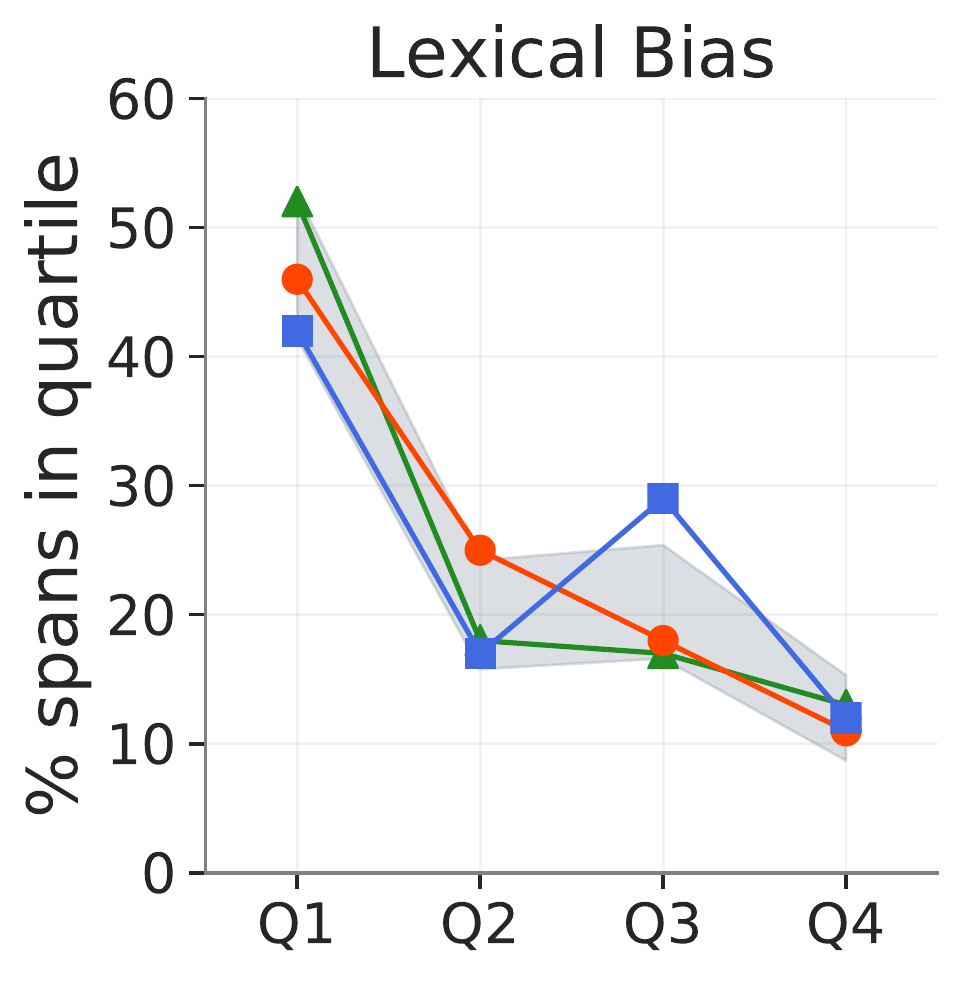}%
\includegraphics[height=5cm]{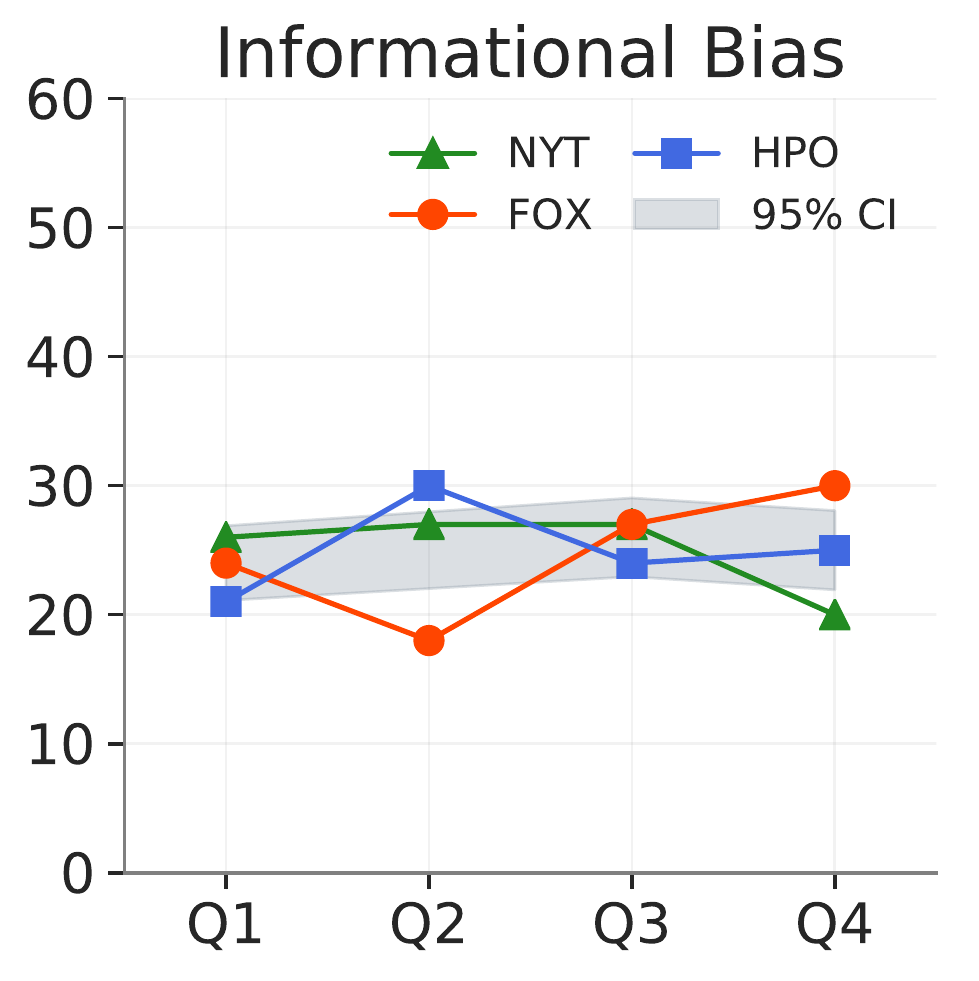}%
}
\vspace{-7mm}
\caption{Distribution of lexical and informational bias spans found in each quartile of an article. The shaded area represents the 95\% confidence interval for the three outlets combined.}
\label{fig:position_trends}
\end{figure}

\subsection{Contrasting the Bias Types}
\noindent \textbf{Informational bias outnumbers lexical bias.} As shown in Table~\ref{tab:descriptive}, the large majority of annotations in \dataset are classified as informational bias. One explanation for its prevalence is that journalists typically make a conscious effort to avoid biased language, but can still introduce informational bias, either intentionally or through negligence.

For both bias types though, negative bias spans are much more pervasive than positive spans, mirroring the well-established paradigm that news media in general focuses on negative events \citep{niven2001bias, patterson1996bad}.

\smallskip
\noindent\textbf{Lexical bias appears early in an article.}
We further study differences in characteristics between lexical and informational annotation spans and find that the two bias types diverge in positional distributions.
Figure~\ref{fig:position_trends} shows that a disproportionate amount of lexical bias is located in the first quartile of articles.
A visual inspection indicates that this may be attributed in part to media sources' attempts to hook readers with inflammatory speech early on (e.g., FOX: ``Paul Ryan stood his ground against a barrage of Biden \textit{grins, guffaws, snickers and interruptions}.'').

In contrast, informational bias is often embedded in context, and therefore can appear at any position in the article. This points to a future direction of bias detection using discourse analysis.

\smallskip
\noindent\textbf{Quotations introduce informational bias.}
We also find that almost half of the informational bias comes from within quotes (48.7\%), highlighting a bias strategy where media sources select opinionated quotes as a subtle proxy for their own opinions (see the second HPO and first NYT annotations in Figure~\ref{fig:example}). 

\subsection{Portrayal of Political Entities}

\begin{figure}
    \centering
% \includegraphics[width=.45\linewidth]{figures/negative_inf.pdf}
% \includegraphics[width=.45\linewidth]{figures/negative_lex.pdf}
% \includegraphics[width=.45\linewidth]{figures/bubble-inf.png}
% \includegraphics[width=.38\linewidth]{figures/bubble-lex.png}
%     \caption{Percentage of bias spans with negative polarity toward targets of known ideology. Number of spans shown by relative size of circles. Ratio of negative spans shown in color. For NYT, 88.66\% of the 247 informational bias spans toward conservatives are negative.}
\resizebox{.95\linewidth}{!}{%
\includegraphics[height=5cm]{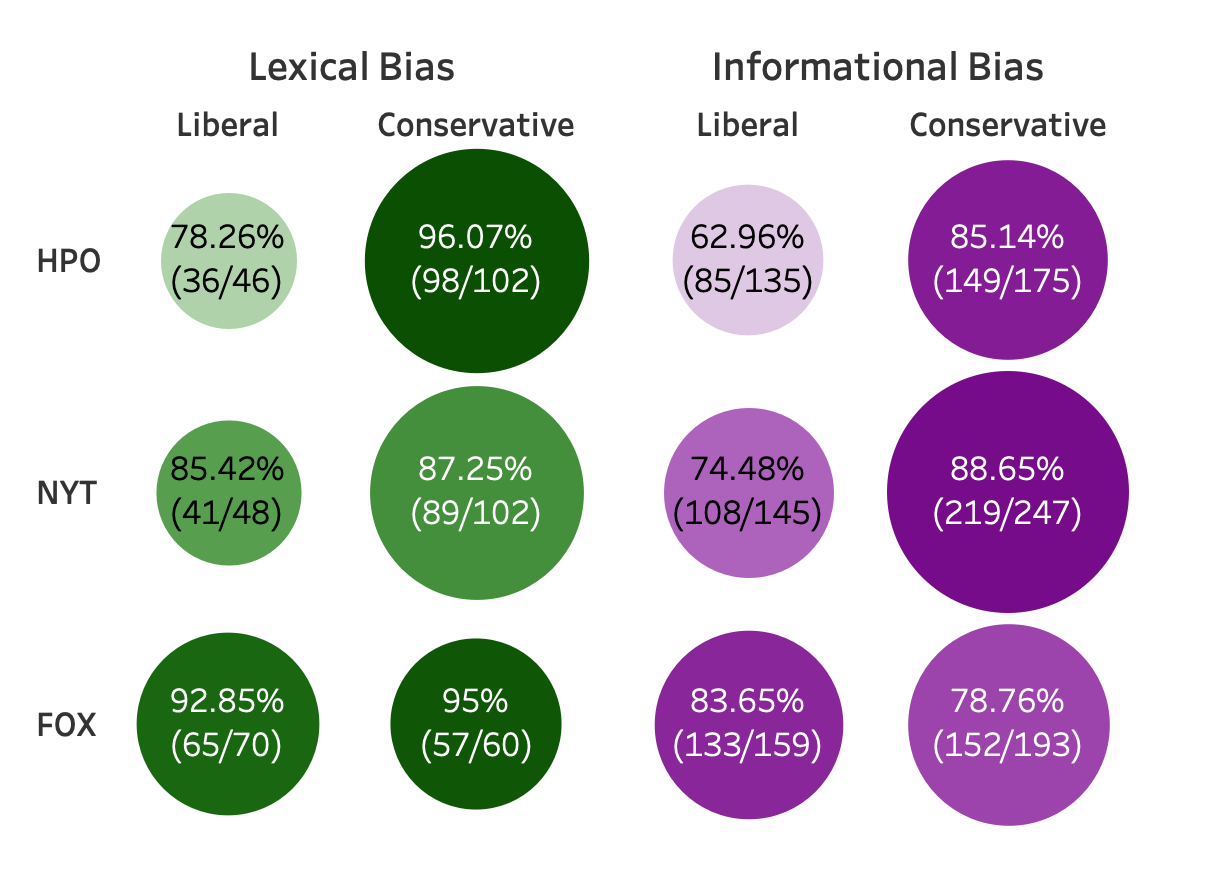}%

}
\vspace{-5mm}
\caption{Percentage of bias spans with negative polarity toward targets of known ideology, grouped by media source, bias type, and target's ideology. For example, in all HPO articles, there are 46 lexical bias spans targeting liberals, 78.26\% of which are negative.
Larger circle means greater number of spans. Darker color corresponds to higher ratio of negative spans.}
% Number of spans shown by relative size of circles. Ratio of negative spans shown in color. For NYT, 88.66\% of the 247 informational bias spans toward conservatives are negative.}
\label{fig:source_heatmap}
\end{figure}

On the document-level, only 17 out of 100 article sets had reversed orderings (i.e. FOX marked as ``more liberal'' or HPO marked as ``more conservative'' within a triplet), confirming the ideological leanings identified in previous studies. 
Here, we utilize \dataset's span-level annotations to gain a more granular picture of how sources covering the same events control the perception of entities.

Concretely, we examine the polarity of bias spans with target entities of known ideology. 
As shown in Figure~\ref{fig:source_heatmap}, for both bias types, the percentage and volume of negative coverage for liberal entities strongly correspond to the ideological leaning of the news outlet. 
Note that though NYT appears to have significantly more informational bias spans against conservatives than HPO, this is because NYT tends to have longer articles than the other two sources (see Table~\ref{tab:descriptive}), and thus naturally results in more annotation spans by raw count.\footnote{The proportion of annotations to article length are similar for all news outlets: one annotation for every 4.1 (for HPO), 4.5 (for FOX), or 4.6 (for NYT) sentences.}

Moreover, the breakdown of lexical bias distinguishes FOX from the other two outlets: it comparatively has more negative bias spans towards liberals and fewer towards conservatives, even though all three outlets have more conservative entities than liberal ones across the 100 triplets (average of 99.0 conservatives, 72.7 liberals).

\section{Experiments on Bias Detection}\label{experiments}
\newcommand{\emphasize}[1]{\textit{#1}}
\begin{table}
\centering
\small
\begin{tabular}{c}
\begin{tabular}{|l r r r|}
\hline
{\bf Sentence-level}& \multicolumn{1}{c}{\bf Precision} & \multicolumn{1}{c}{\bf Recall} & \multicolumn{1}{c|}{\bf F1}\\
\hline
\multicolumn{4}{|l|}{\textit{Lexical Bias}}\\ 
% BERT fine-tuning & 29.1 $\pm$ 5.9 & 38.6 $\pm$ 7.2 & 31.5 $\pm$ 4.7\\
% BERT fine-tuning & 29.13 $\pm$ 5.9 & 38.57 $\pm$ 7.2 & 31.49 $\pm$ 4.7\\
BERT fine-tuning & 29.13 & 38.57 & 31.49 \\
%\multicolumn{1}{|r}{+ context} & 27.83 & 45.77 & 32.61 \\
\hdashline
\multicolumn{4}{|l|}{\textit{Informational Bias}}\\ 
TF-IDF & 25.81 & 26.23 & 26.02 \\
% BERT fine-tuning & 43.9 $\pm$ 3.8 & 42.9 $\pm$ 2.9 & 43.3 $\pm$ 1.9\\
% BERT fine-tuning & 43.87 $\pm$ 3.8 & 42.91 $\pm$ 2.9 & 43.27 $\pm$ 1.9\\
BERT fine-tuning & 43.87 & 42.91 & 43.27 \\
%\multicolumn{1}{|r}{+ context} & 37.90 & 45.12 & 40.58 \\
\hline
\end{tabular}\\
\\
\begin{tabular}{|l r r r|}
\hline
{\bf Token-level}& \multicolumn{1}{c}{\bf Precision} & \multicolumn{1}{c}{\bf Recall} & \multicolumn{1}{c|}{\bf F1}\\
\hline
\multicolumn{4}{|l|}{\textit{Lexical Bias}}\\ 
Polarity lexicon & 8.00 & 0.17 & 0.33 \\
Subjectivity lexicon & 28.00 & 0.65 & 1.28 \\
% BERT fine-tuning & 25.6 $\pm$ 4.4 & 29.3 $\pm$ 4.6 & 26.0 $\pm$ 3.6 \\
% BERT fine-tuning & 25.60 $\pm$ 4.4 & 29.32 $\pm$ 4.6 & 25.98 $\pm$ 3.6 \\
BERT fine-tuning & 25.60 & 29.32 & 25.98 \\
\hdashline
\multicolumn{4}{|l|}{\textit{Informational Bias}}\\ 
% BERT fine-tuning & 25.6 $\pm$ 3.8 & 14.8 $\pm$ 2.5 & 18.7 $\pm$ 3.0\\
% BERT fine-tuning & 25.56 $\pm$ 3.8 & 14.78 $\pm$ 2.5 & 18.71 $\pm$ 3.0\\
BERT fine-tuning & 25.56 & 14.78 & 18.71\\
%\multicolumn{1}{|r}{+ context} & 7.94 & 5.38 & 6.41 \\
\hdashline
% \hline
\multicolumn{4}{|l|}{\textit{Sentence-to-Token pipeline}}\\ 
Lexical bias & 12.00 & 13.64 & 12.77 \\
Informational bias & 9.52 & 5.08 & 6.63 \\
\hline
\end{tabular}
% \\
% \\
% \begin{tabular}{|l r r r|}
% \hline
% \bf Sentence/Token Pipeline & P & R & F\\ 
% \hline
% Informational & 9.52 & 5.08 & 6.63 \\
% Lexical & 12.00 & 13.64 & 12.77 \\
% \hline
% \end{tabular}
\end{tabular}
\vspace{-2mm}
\caption{Sentence classification (top) and sequence tagging (bottom) results on lexical and informational bias prediction. For the BERT fine-tuning models, the mean from 10-fold cross validation is shown. The minimum standard deviation from cross validation for all BERT models is 3.36, the maximum is 12.44.}\label{tab:bert}
\end{table}

%\multicolumn{4}{|l|}{\textit{Lexical Bias}}\\ 
% Sent classifier & 21.74 & 50.00 & 30.30 \\
% \multicolumn{1}{|r}{+ context} & 21.74 & 41.67 & 28.57 \\
% \hdashline
% Polarity baseline & 8.00 & 0.17 & 0.33 \\
% Subjectivity baseline & 28.00 & 0.65 & 1.28 \\
% MPQA baseline & 32.00 & 0.42 & 0.83 \\
% Seq tagger & 36.00 & 28.12 & 31.58 \\
% Pipeline & 16.00 & 26.67 & 20.00 \\
% \hline
% \multicolumn{4}{|l|}{\textit{Informational Bias}}\\ 
% TF-IDF baseline & 25.81 & 26.23 & 26.02 \\
% Sent classifier & 27.42 & 56.67 & 36.95 \\
% \multicolumn{1}{|r}{+ context} & 37.10 & 46.94 & 41.44 \\
% \hdashline
% Seq tagger & 19.05 & 11.01 & 13.95 \\
% Pipeline & 3.17 & 3.28 & 3.23 \\
% \multicolumn{1}{|r}{+ context} & 9.52 & 6.67 & 7.84 \\
We study the bias prediction problem on \dataset as a binary classification task (i.e., whether or not a sentence contains bias) and as a BIO sequence tagging task (i.e., tagging the bias spans in one sentence at the token-level). We benchmark the performance with rule-based classifiers and the popular BERT model \citep{devlin2019bert} fine-tuned on informational and lexical bias spans separately.

\smallskip
\noindent\textbf{Training Details.} We utilize the pre-trained BERT-Base model and use the ``Cased'' version to account for named entities, which are important for bias detection. We run BERT on individual sentences\footnote{BERT's maximum input length is 512 tokens, which is shorter than most articles in \dataset. We thus treat sentences as passages, rather than using text of fixed length.}
and perform stratified 10-fold cross validation. The validation set is used to determine when to stop training and a held out test set is used for the final evaluation of each fold. For the sentence-level classifiers, both our informational and lexical models use 6,819 sentences for training, 758 for validation, and 400 for testing. 

Due to the sparsity of our data, we train and test our token-level models only on sentences containing bias spans of the relevant bias type. Our informational and lexical bias sequence taggers use a train/val/test split of 1,043/116/62 sentences and 383/42/23 sentences respectively. Results are shown in Table~\ref{tab:bert}.

\smallskip
\noindent\textbf{Sentence-level Classifier.} The fine-tuned BERT is better at predicting informational bias than lexical bias, likely because informational bias is better captured by sentence-level context.
As a baseline, we select the 4 sentences\footnote{\dataset averages 4.1 informational bias spans per article.} in each article with the lowest average TF-IDF token scores as containing informational bias.
The intuition is that sentences {with different content than the rest of the article are more likely to contain extraneous information} that the author chose to include to frame the story in a certain way. 
We find that this simple baseline performs relatively well considering the difficulty of the task, indicating the importance of explicitly modeling context.
Future work may consider leveraging context in the entire article or articles on the same story by other media.

\smallskip
\noindent\textbf{Token-level Classifier.} 
From Table~\ref{tab:bert}, we see that the BERT lexical sequence tagger produces better recall and F1 than the informational tagger, highlighting the additional difficulty of accurately identifying spans of informational bias. 
We also use the polarity and subjectivity lexicons from the MPQA website \citep{wilson2005recognizing,choi2014+} as a simple baseline for lexical bias tagging and find that these word-level cues, though widely used in prior sentiment analysis studies, are insufficient to fully capture lexical bias. 

In order to evaluate token-level prediction on the larger original test set, we conduct a pipeline experiment with the fine-tuned BERT models where sentences predicted as containing bias by the best sentence-level classifier from cross validation are tagged by the best token-level model. The results reaffirm our hypothesis that while both tasks are extremely difficult, informational bias is more challenging to detect.

\section{Conclusion}\label{conclusion}
We presented a novel study on the effects of informational bias in news reporting from three major media outlets of different political ideology. 
Analysis of our annotated dataset, \dataset, showed the prevalence of informational bias in news articles when compared to lexical bias, and demonstrated \dataset's utility as a fine-grained indicator of how media outlets cover political figures. 
An experiment on bias prediction illustrated the importance of context when detecting informational bias and revealed future research directions.

\section*{Acknowledgements}
This research is supported in part by National Science Foundation through Grant IIS-1813341. 
We thank Philip Resnik, Nick Beauchamp, and Donghee Jo for their valuable suggestions on various aspects of this work. 
We are also grateful to the anonymous reviewers for their comments.

\bibliography{references}
\bibliographystyle{acl_natbib}

\appendix

\section{Sample Annotations}

On the right, several sample annotations from the \dataset dataset illustrate some aspects of our annotation schema and highlight characteristics of informational bias.

\medskip
\noindent \textbf{Indirect Bias.} Though not as prevalent as bias spans with direct aim, indirect aim is nevertheless important to study because readers may find it more difficult to detect bias consciously when it does not directly implicate the main entity. Indirect bias can be aimed through an intermediary ally or opponent, or may be based on contextual information. In each case, the sentiment towards the intermediary entity alters sentiment toward the main target entity.
%is transferred to

\begin{figure}[t!]
\centering
% \small
\fontsize{8.5}{10.83}
\fontfamily{phv}\selectfont
\begin{subfigure}[t]{.95\linewidth}
\begin{tabular}{|p{\linewidth}|}
\hline
\textbf{Main Event:} Trump reverses decision to allow import of elephant trophies\\
\textbf{Main Entity:} Donald Trump \\
\hdashline
\textbf{NYT:} 
%President Trump on Friday reversed the government's decision to start allowing hunters to import trophies of elephants that were killed in two African countries, pending a further review. ... 
On social media, photos were being shared of Mr. Trump's two elder sons hunting on safari in Zimbabwe, [\textit{including one photo that showed \underline{Donald Trump Jr.} with a severed elephant tail in one hand and a knife in the other.}]$_{\textcolor{red}{\text{Trump}}}$\\
\hline
\end{tabular}
\caption{Indirect negative informational bias against Donald Trump, using the intermediary entity Donald Trump Jr.}
\label{fig:samples_elephants}
\vspace{4mm}
\end{subfigure}

\begin{subfigure}[t]{.95\linewidth}
\begin{tabular}{|p{\linewidth}|}
\hline
\textbf{Main Event:} Trump declares national emergency over border wall\\
\textbf{Main Entity:} Donald Trump\\
\hdashline
\textbf{HPO:} 
%President Donald Trump on Friday declared a national emergency at the southern U.S. border... 
[\textit{Since 2014, a high proportion of those crossing have been Central American \underline{children and families} seeking to make humanitarian claims such as asylum.}]$_{\textcolor{red}{\text{Trump}}}$\\
\hdashline
\textbf{FOX:} President Trump said Friday he is declaring a national emergency on the southern border ... [\textit{despite his criticisms of former President Barack Obama for using executive action.}]$_{\textcolor{red}{\text{Trump}}}$\\
\hdashline
\textbf{NYT:} 
Mr. Trump's announcement came during a freewheeling, 50-minute appearance ... [\textit{The president again suggested that he should win the Nobel Peace Prize, and he reviewed which conservative commentators had been supportive of him, while dismissing Ann Coulter, who has not.}]$_{\textcolor{red}{\text{Trump}}}$\\
\hline
\end{tabular}
\caption{Example annotations showing negative informational bias from all three media sources for one article triplet.}
\label{fig:samples_outlets}
\vspace{4mm}
\end{subfigure}

\begin{subfigure}[t]{.95\linewidth}
\begin{tabular}{|p{\linewidth}|}
\hline
\textbf{Main Event:} Raul Labrador challenges Kevin McCarthy for House majority leadership\\
\textbf{Main Entities:} Raul Labrador, Kevin McCarthy\\
\hdashline
\textbf{HPO:} [\textit{Labrador is an ambitious, sometimes savvy politician.}]$_{\text{{\textcolor{red}{Labrador}}}}$ He is in Idaho this weekend chairing the state GOP convention.\\
\hline
\end{tabular}
\caption{Example annotation of positive informational bias.}
\label{fig:samples_obama}
\end{subfigure}

% \begin{subfigure}[t]{\linewidth}
% \begin{tabular}{|p{\linewidth}|}
% \hline
% \textbf{Main Event:} Hillary Clinton suggests Obama should listen to his allies on trade\\
% \hdashline
% \textbf{HPO:} House Democrats rejected Obama's trade agenda Friday by blocking [\textit{a measure that would have granted him the power to fast-track sweeping, secretive international agreements through Congress.}]$_{\textrm{{\textcolor{blue}{Obama}}}}$\\
% \hline
% \end{tabular}
% \caption{Annotation showing contextual information in a neutral tone that appears objective, but is informational bias.}
% \label{fig:samples_obama}
% \end{subfigure}

\caption{Excerpts showing different types of informational bias, annotated in italics. The target of the negative bias is noted at the end of each span. Underlined entities are intermediary targets in indirect bias spans.}\label{fig:examples}
\end{figure}
Figure~\ref{fig:samples_elephants} shows an example of indirect bias where Donald Trump is negatively targeted via the negative framing of an ally, Donald Trump Jr. Readers are required to know the relationship between the two men in order to notice the bias, and the information itself would be irrelevant to the article were it not for their relationship.
%The information is distantly related to the main event and an understanding of the relationship between Trump and his ally is required to identify the bias. 

% \input{figures/example_wall.tex}
The span from HPO in Figure~\ref{fig:samples_outlets} shows an indirect bias span where contextual information 
%about children and families, 
unconnected to the rest of the article reflects negatively on Trump without mentioning him in the text. It requires several leaps in logical thinking: children and families seeking asylum are sympathetic :: turning them away is bad :: Trump wants a border wall :: Trump is framed negatively. This type of informational bias is difficult to detect algorithmically as there is no mention of Trump, the target main entity. 

\medskip
\noindent \textbf{Informational Bias Strategies.}
Inspecting the informational bias spans in our dataset reveals several trends and strategies that journalists tend to use.
%Figure~\ref{fig:samples_outlets} shows informational bias from all three media sources for one triplet. 
The examples from FOX and NYT in Figure~\ref{fig:samples_outlets} show the strategy where objective but tangential information frames the target in a negative light given the context of the article.
The example from FOX uses nonessential background information to imply Trump is hypocritical, and the NYT example includes a detail peripheral to the main event that portrays Trump as rambling.

Figure~\ref{fig:samples_obama} is an example of subtle informational bias where the author's opinion masquerades as fact. The writing is in a neutral tone and appears objective, but it is actually the author's perception of the situation and uncovers their bias towards the topic. The span is categorized as informational bias rather than lexical because there is no way to rephrase or remove parts of the sentence without changing the overall meaning.
This span is also an example of the rarer positive bias span.

\section{Data Collection}

\dataset contains 100 triplets of articles, each with 3 articles about the same main event from the New York Times (NYT), Fox News (FOX), and the Huffington Post (HPO). According to \citet{budak2016fair}, FOX is considered strongly right leaning, NYT slightly left leaning, and HPO strongly left leaning. %10 triplets are selected for each year from 2010 to 2019. 
As an initial annotation set, 16 triplets of highly visible, polarizing events were directly selected from the media source websites by our annotators. 

The remaining triplets were aligned algorithmically from the Common Crawl corpus.\footnote{\url{http://commoncrawl.org}} Articles with less than 200 words or more than 1,000 words were filtered out, and only political, non-editorial articles published within 3 days of each other were considered. Article similarity was calculated using the cosine similarity of the TF-IDF vectors of each article's title combined with its first 5 sentences. For each FOX article, the most similar NYT article was found, then the most similar HPO article was found using this pair. An annotator manually selected the final triplets from this list of automatically aligned triplets. 

Main event and entities were manually annotated for each article by one annotator. Articles in a triplet share the same main event, which the annotator produced after reading the leads of the three articles. Main entities sometimes differ across the triplet, as stories about the same event can emphasize different characters, but at least one main entity is consistent across each triplet. A single article contains an average of 2.04 main entities and at most five main entities. 

During the annotation process, the order of articles is randomized within each triplet and annotators are not aware of the media source of the article. The entire dataset was annotated by three unique annotators.

%Table~\ref{tab:descriptive} shows the statistics of our dataset separated by media source. %Table~\ref{tab:inf_lex_stats} shows general statistics separated by bias type. 
%NYT articles tend to be longer than articles from the other sources, thus naturally resulting in more annotation spans by raw count. However, we find that the proportion of annotations per sentence across all outlets is approximately the same.

% \section{Dataset Description}
% \input{further_analysis.tex}

%\blindtext
%\begin{figure}[htp]
%\minipage{0.32\textwidth}
% \includegraphics[width=\linewidth]{figures/tool_none_selected.png}
%  \caption{}\label{fig:awesome_image1}
%\endminipage\hfill
%\minipage{0.32\textwidth}
%  \includegraphics[width=\linewidth]{figures/tool_annotation_form.png}
%  \caption{}\label{fig:awesome_image2}
%\endminipage\hfill
%\minipage{0.32\textwidth}%
%  \includegraphics[width=\linewidth]{figures/tool_article_form.png}
%  \caption{}\label{fig:awesome_image3}
%\endminipage
%\label{fig:tool}
%\end{figure}
%\blindtext

\begin{figure*}[t]
\begin{subfigure}[t]{0.33\textwidth}
\includegraphics[width=\textwidth]{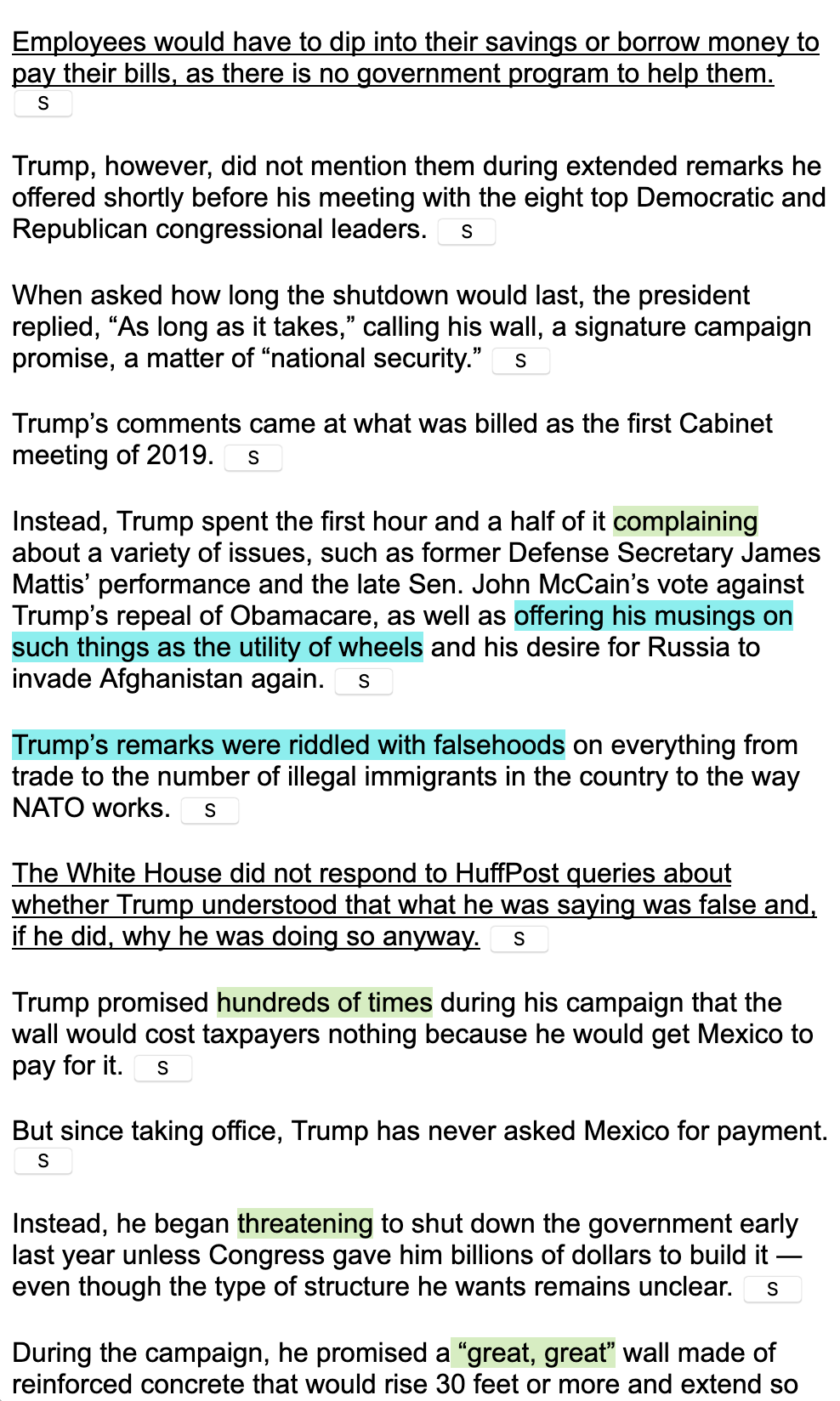}
\subcaption{Tool with loaded annotations.\\Informational bias spans are shown\\in blue, lexical bias spans are shown\\in green.}
\label{fig:tool_none_selected}
\end{subfigure}
\begin{subfigure}[t]{0.33\textwidth}
\includegraphics[width=\textwidth]{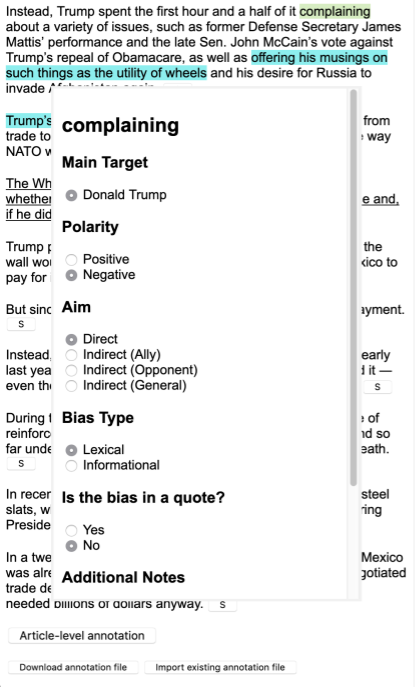}
\subcaption{Tool with sentence-level annotation form.}
\label{fig:tool_span_form}
\end{subfigure}
\begin{subfigure}[t]{0.33\textwidth}
\includegraphics[width=\textwidth]{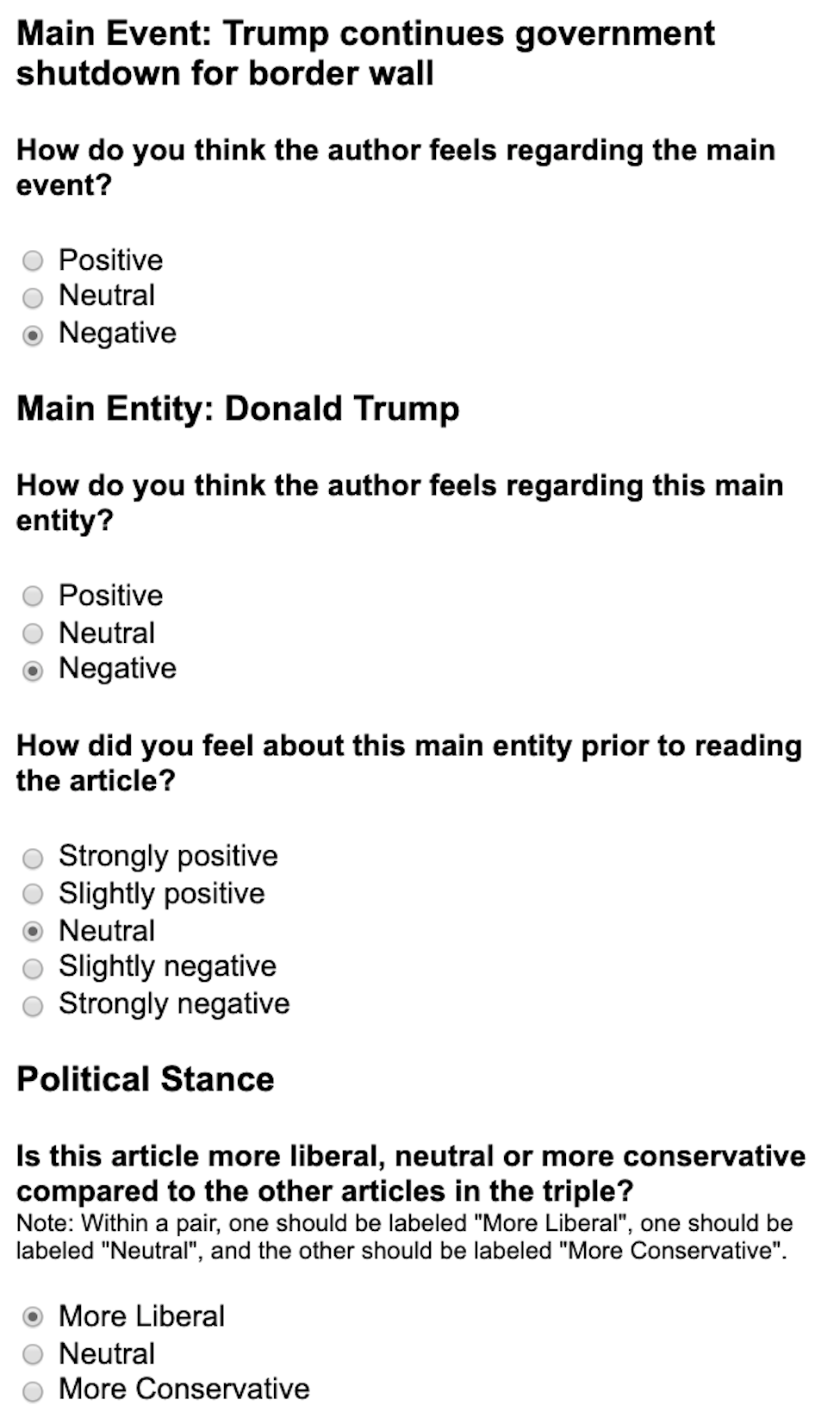}
\subcaption{Document-level annotation form.}
\label{fig:tool_article_form}
\end{subfigure}
\caption{Our Javascript annotation tool at various steps.}
            % \includegraphics[width=.3\textwidth]{figures/tool_annotation_form.png}\hfill
            % \includegraphics[width=.3\textwidth]{figures/tool_article_form.png}
            % \caption{Image A.}
            % \caption{Image B.}
            % \caption{Image C.}
\end{figure*}

\section{Inter-annotator Agreement}
\begin{table}[t]
\small
\centering
% \begin{tabular}{|c|c|c|c|}
% \toprule
% \hline
% \toprule
%       &               A+B &             A+C &             B+C \\
% Match style & P {} {} {} {} {} R {} {} {} {} {} F & P {} {} {} {} {} R {} {} {} {} {} F & P {} {} {} {} {} R {} {} {} {} {} F \\
% \hline
% \midrule
% lenient &  0.12 0.16 0.14 &  0.11 0.13 0.12 &  0.15 0.15 0.15 \\
% exact  &  0.11 0.14 0.12 &   0.09 0.1 0.09 &           0 0 0 \\
% \bottomrule
% \hline
% \end{tabular}
\begin{tabular}{|c| c c c | c c c |}
\hline
 & \multicolumn{3}{c|}{\textbf{Exact Matching}}& \multicolumn{3}{c|}{\textbf{Lenient Matching}}\\
& Prec. & Rec. & F1 & Prec. & Rec. & F1 \\
\hline
\multicolumn{4}{|l}{\textit{Lexical Bias}}&&&\\
A + B &  11.04 & 14.17 & 12.41 & 12.34 & 15.83 & 13.87 \\
A + C & \hfill8.57 & \hfill9.76 & \hfill9.13  &  11.43 & 13.01 & 12.17 \\
B + C  &   -- & -- & -- &  15.38 & 15.38 & 15.38 \\
\hdashline
\multicolumn{4}{|l}{\textit{Informational Bias}}&&&\\
A + B & 19.90 & 17.22 & 18.46 & 39.80 & 34.33 & 36.92 \\
A + C &  19.47 & 22.05 & 20.68 &  34.40 & 38.97 & 36.54 \\
B + C  &  15.29 & 10.83 & 12.68 &  32.94 & 23.33 & 27.32 \\
% A + GS & 0.90 & 0.62 & 0.73 &0.92 & 0.63 & 0.75  \\
% B + GS & 0.27 & 0.47 & 0.34 & 0.34 & 0.59 & 0.43 \\
% C + GS &  0.60 & 0.33 & 0.43  & 0.70 & 0.39 & 0.50 \\
% \hline
% \hdashline
% A + GS & 0.89 & 0.67 & 0.76 &  0.94 & 0.70 & 0.80  \\
% B + GS &  0.44 & 0.53 & 0.48  & 0.62 & 0.75 & 0.68 \\
% C + GS & 0.53 & 0.34 & 0.42  & 0.78 & 0.50 & 0.61 \\
\hline
\end{tabular}
\caption{Inter-annotator span agreement for lexical and informational bias. Dashes indicate that there were no exact matching lexical text spans between annotators B and C.}
\label{tab:iaa_span_lex}
\end{table}

% \input{tables/iaa_span_inf.tex}
% \begin{table*}[t!]
% \centering
% \small
% % \begin{tabular}{|c|c|c|c|}
% % \toprule
% % \hline
% % Dimension & A+B & A+C & B+C \\
% % \hline
% % \midrule
% % polarity  &  0.84 &  0.75 &  0.92 \\
% % aim       &  0.54 &  0.12 &   0.0 \\
% % target    &  0.93 &  0.88 &  0.96 \\
% % \bottomrule
% % \hline
% % \end{tabular}
% \begin{tabular}{|c|c||c c c||c c c|}
% \hline
% & & \multicolumn{3}{c||}{Dimensions (Cohen's $\kappa$)}
% & \multicolumn{3}{c|}{Dimensions (\% Agreement)}\\
% & \# Triples & Target &  Polarity & Aim & Target &  Polarity & Aim\\
% \hline
% A + B & 41 & 0.93 & 0.84 &  0.54 & 0.89 & 0.95 & 0.90 \\
% A + C & 46 & 0.88 & 0.75 &  0.12 & 0.94 & 0.96 & 0.94  \\
% B + C & 13 & 0.96 & 0.92 &  -- & 0.97 & 0.97 & 0.97  \\
% \hline
% \end{tabular}
% \caption{Number of triples resolved by each annotator pairing, Cohen's $\kappa$, and percent agreement for inter-annotator agreements on auxiliary dimensions for overlapping spans.}
% \label{tab:iaa_attributes}
% \end{table*}

\begin{table}[t!]
\centering
\small
\begin{tabular}{|c|c||c c c|}
\hline
& & \multicolumn{3}{c|}{\textbf{Dimensions} (Cohen's $\kappa$ / \% Agr.)}\\
& \# Res. & Target &  Polarity & Aim \\
\hline
% A + B & 41 & 0.93 / 0.89 & 0.84 / 0.95 &  0.54 / 0.90\\
% A + C & 46 & 0.88 / 0.94 & 0.75 / 0.96 &  0.12 / 0.94\\
% B + C & 13 & 0.96 / 0.97 & 0.92 / 0.97 &  \multicolumn{1}{r|}{\hfill--\hfill/ 0.97} \\
A + B & 123 & 0.93 / 93.7 & 0.84 / 96.3 &  0.12 / 93.7\\
A + C & 138 & 0.88 / 89.5 & 0.75 / 95.0 &  0.54 / 89.9\\
B + C & 39 & 0.96 / 96.9 & 0.92 / 96.9 &  \multicolumn{1}{r|}{\hfill--\hfill/ 96.9} \\

\hline
\end{tabular}
\caption{Number of articles resolved by each annotator pairing, along with Cohen's $\kappa$ and percent agreement for IAA on auxiliary dimensions for overlapping spans.}
\label{tab:iaa_attributes}
\end{table}

Our study of inter-annotator agreement consists of two parts: the agreement of the text spans selected and the agreement on the dimensions within each annotation span. 
To find text span agreement, a similar method to \citet{toprak2010sentence} is used in which precision, recall, and F1 are calculated between two annotators using the agreement metric from \citet{wiebe2005annotating}, treating one annotator's spans as the gold standard and the other annotator's spans as the system. Results are calculated for \textit{exact matching}, where the text spans must overlap exactly to be considered correct, and \textit{lenient matching}, where text spans with any overlaps are considered correct \citep{somasundaran2008discourse}.
%using the \textit{exact} and \textit{lenient} text span matching styles used by \citet{somasundaran2008discourse}. \textit{Exact} measures spans which overlap exactly, and \textit{lenient} measures spans which overlap to some degree but which may have differing text boundaries. 

Table~\ref{tab:iaa_span_lex} shows that span agreement is higher for spans of informational bias than for spans of lexical bias due to the sparsity of lexical bias in our dataset (see Table 1 in the main paper).

%We calculate span agreement separately for the two bias types, showing in Table~\ref{tab:iaa_span_lex} that span agreement is not only significantly higher in lenient matching than in exact matching, but is also higher for spans of informational bias than for spans of lexical bias due to the sparsity of lexical bias in our dataset.

Dimension agreement is reported in Table~\ref{tab:iaa_attributes} only for lenient matching spans, as the results are not significantly different from that of exact matching spans. Cohen's $\kappa$ is used to measure attribute agreement for target, polarity, and aim, and we find high levels of agreement for both polarity and target. Because of the metric's sensitivity to class imbalance, Cohen's $\kappa$ is impractical for measuring the agreement on aim for one annotator pairing (B + C), which had fewer article triplets to resolve and nearly all overlapping lexical annotations were marked as \textit{direct} (31 / 32 spans). To account for this imbalance, the percent agreement for all attributes is also included in Table~\ref{tab:iaa_attributes}.

%use the simple metric of percent agreement and find the agreement for the aim attribute averaged across all annotators is 0.94.

% \section{BERT Training Details}
% \input{bert.tex}

\section{Javascript Annotation Tool}
% figures currently in iaa.tex for positioning
% \input{figures/tool.tex}

A Javascript based tool\footnote{\url{https://github.com/marshallwhiteorg/emnlp19-media-bias}} was developed to annotate our dataset. Annotations created in the tool can be downloaded in JSON format and analyzed or imported at a later date. Users can highlight spans of text or select an entire sentence, then answer dimensional questions (see Figure~\ref{fig:tool_span_form}). Users can also answer document-level questions (see Figure~\ref{fig:tool_article_form}).
Figure~\ref{fig:tool_none_selected} shows the tool after annotations have been made, where blue spans are informational bias and green spans are lexical bias. In order to alleviate eye strain, annotations of the entire sentence are shown underlined rather than highlighted.
%Underlined text means the entire sentence has been annotated as biased. 
%Figure~\ref{fig:tool_span_form} shows the form an annotator uses to record their attribute-level answers, and Figure~\ref{fig:tool_article_form} shows the form for answering article-level questions.

\end{document}